\documentclass[preprint,12pt]{elsarticle}

\usepackage{amssymb}
\usepackage{multirow} 
\usepackage{tikz,pgfplots} 
\usepackage{siunitx} 
\usepackage{graphicx}
\usepackage{subcaption}
\usepackage{hyperref} 

\usetikzlibrary{bayesnet} 

\journal{Renewable Energy}

\begin{document}

\begin{frontmatter}






\title{Multitask learning for improved scour detection: A dynamic wave tank study}


\author[The University of Sheffield]{Simon M. Brealy\corref{cor1}}
\ead{sbrealy1@sheffield.ac.uk}

\affiliation[The University of Sheffield]{organization={Dynamics Research Group},
            addressline={Department of Mechanical Engineering, The University of Sheffield, Mappin Street}, 
            city={Sheffield},
            postcode={S1 3JD}, 
            country={UK}}

\author[The University of Sheffield]{Aidan J. Hughes}
\author[The University of Sheffield]{Tina A. Dardeno}
\author[The University of Glasgow]{Lawrence A. Bull}
\author[The University of Sheffield]{Robin S. Mills}
\author[The University of Sheffield]{Nikolaos Dervilis}
\author[The University of Sheffield]{Keith Worden}

\affiliation[The University of Glasgow]{organization={School of Mathematics and Statistics},
            addressline={University of Glasgow}, 
            city={Glasgow},
            postcode={G12 8TA}, 
            country={Scotland}}

\cortext[cor1]{Corresponding author}

\begin{abstract}
Population-based structural health monitoring (PBSHM), aims to share information between members of a population. An offshore wind (OW) farm could be considered as a population of nominally-identical wind-turbine structures. However, benign variations exist among members, such as geometry, sea-bed conditions and temperature differences. These factors could influence structural properties and therefore the dynamic response, making it more difficult to detect structural problems via traditional SHM techniques.

This paper explores the use of a Bayesian hierarchical model as a means of multitask learning, to infer foundation stiffness distribution parameters at both population and local levels. To do this, observations of natural frequency from populations of structures were first generated from both numerical and experimental models. These observations were then used in a partially-pooled Bayesian hierarchical model in tandem with surrogate FE models of the structures to infer foundation stiffness parameters. Finally, it is demonstrated how the learned parameters may be used as a basis to perform more robust anomaly detection (as compared to a no-pooling approach) e.g.\ as a result of scour.
\end{abstract}


\begin{highlights}
\item Bayesian hierarchical models help reduce uncertainty of foundation model parameters in populations of wind-turbines
\item Reduced foundation parameter uncertainty aids detection of anomalies in dynamic behaviour during operation
\item Future design of turbines may also be improved through reducing the likelihood and severity of fatigue damage 
\end{highlights}

\begin{keyword}
Bayesian hierarchical modelling \sep population-based SHM \sep wind-turbine foundations \sep scour




\end{keyword}

\end{frontmatter}


\section{Introduction}
\label{sec:intro}


\noindent The design of offshore wind (OW) turbine towers and foundations is driven by fatigue and extreme loading concerns \citep{noauthor_guide_2019}. To minimise fatigue damage, the resonance frequencies of the dynamic structure must be avoided, to prevent amplification of the response to external forcing \citep{andersen_natural_2012}. These forces predominantly come in the form of wind and wave loading in the natural environment, and forces from the rotation of the blades, at both the rotational speed (1P) and three times the rotational speed (3P). Excitation from the 1P frequency is associated with mass imbalance in the blades, while the 3P frequency is associated with the blade-passing frequency of the tower, causing a shadow effect \citep{bhattacharya_physical_2021}. 

Typically, the first natural frequencies of the structures are designed to lie between the 1P and 3P frequencies, as a compromise between steering clear of environmental loading, and the cost of additional structural material for stiffness \citep{bhattacharya_challenges_2014}. Careful consideration must, therefore, come in the design stage of wind-turbine foundations, to avoid resonance frequencies coinciding with the frequencies of the sources of excitation.

An OW farm could be considered as a homogeneous population of wind-turbine structures \citep{bull_foundations_2021}. However, from the point of view of their dynamic response, variations exist between members of the population; these include possible variations in the soil profile across the wind farm, influencing the effective stiffness of the foundation support. Furthermore, more benign environmental variations also exist, such as in temperature, wave and wind conditions or physical geometry from manufacturing tolerances. According to a study by Sørum et al.~\citep{sorum_fatigue_2022}, dynamic response uncertainties related to wind conditions dominate in the tower top, while uncertainties in the wave and soil models dominate in the tower base and monopile.

Given the relatively-narrow frequency band in which the first natural frequency should lie, minimising uncertainty in the modelling of dynamic response is important to reduce the likelihood of fatigue damage in operation. Furthermore, from the point of view of performing SHM, any uncertainties that do remain in the expected dynamic response may make identification of damage that influences the structural properties more challenging. For example, a significant problem in the operation of wind farms is the scouring of sediment around monopile foundations at the interface between seawater and sea bed; this in effect reduces the embedded depth of the monopile, which, in turn, reduces the stiffness of the support \citep{prendergast_investigation_2013, prendergast_investigation_2015}. As a consequence, the dynamic response of the entire structure is affected. However, if the reduction in natural frequency is within the expected variance in the dynamic response of an individual structure, it may go undetected.

The current work uses multitask learning, which is a method to improve inferences by learning many tasks at once; a natural way to do this is by adopting a Bayesian hierarchical modelling approach (which is also previously used in the application of PBSHM \citep{bull_hierarchical_2022, dardeno_hierarchical_2024}), to develop population and individual turbine-level distributions for the natural frequency of the first bending mode of wind-turbine structures, which share a level of information between each other. Observations of the natural frequencies are generated both numerically - via a Finite-Element (FE) model, and experimentally - via testing in a wave tank - the details of which are described in detail in sections \ref{sec:fe_approach} and \ref{sec:exp_approach}. Posterior distributions describing the uncertainty of the stiffness of the foundations are then recovered, using the observations of natural frequency with the Bayesian hierarchical model (described in Section \ref{sec:bayesian_hierarchical_model}) and surrogate forms of FE models of the structures. Finally, an example showing how these ideas could be used for anomaly detection (such as detecting scour), is shown in Section \ref{sec:scour}.

\section{Methodology}

\noindent Two approaches were used to generate natural frequency data for the Bayesian hierarchical model. The first approach which involved the generation of data via a Finite Element (FE) model of the NREL 5MW reference turbine \citep{jonkman_definition_2009}, was used for validation purposes. The second approach used physical scale models of wind-turbine structures, subject to excitation from waves in a wave tank. The details of these data generation approaches are described in the following sections.

\subsection{FE modelling approach}\label{sec:fe_approach}
\noindent An FE model was constructed to model the dynamic response of a wind-turbine structure, with geometry and material properties based on the widely studied NREL 5MW reference turbine \citep{jonkman_definition_2009}. A tubular construction for the monopile and tower was used with Timoshenko beam elements \citep{davis_timoshenko_1972}, and a lumped mass placed at the top to represent the mass of the nacelle, rotor and blades. For simplicity, the transition piece was not modelled, which is normally used to mate the monopile to the tower. Instead, the cross-sections of the monopile and tower were joined and effectively modelled as a single beam.

\begin{figure}[h] 
    \centering{\includegraphics[width=0.7\columnwidth]{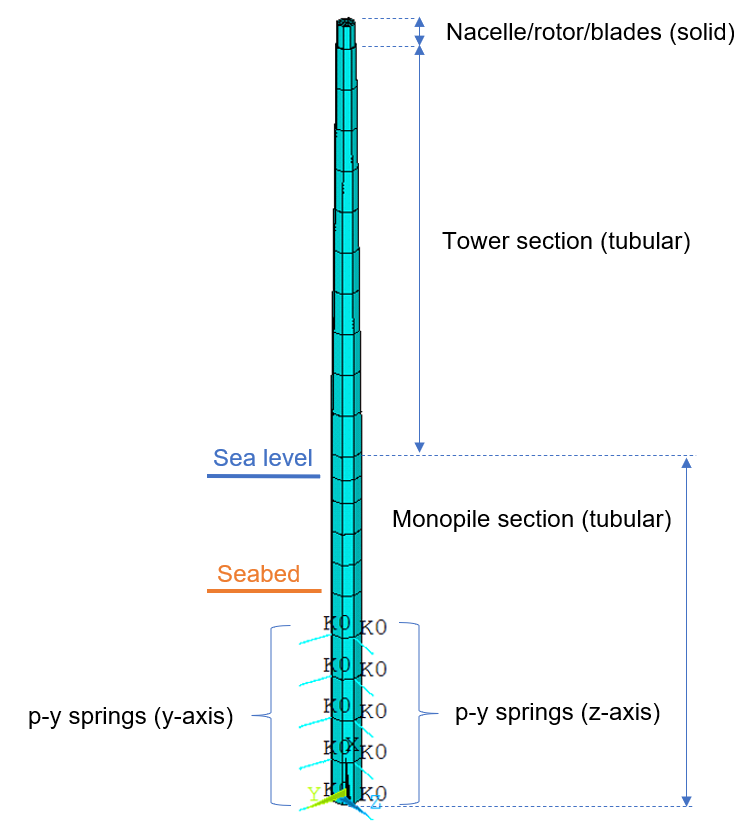}}
      \caption{FE model construction in relation to the offshore environment.}
  \label{fig:fe_model}%
  \end{figure}

To model the soil-structure interaction (SSI), a Winkler foundation was adopted, using a series of springs to resist lateral movement (known as \textit{p-y} springs) \citep{winkler_lehre_1868}. For simplicity, the base of the foundation was pinned to restrict movement in the axial direction; this is justified as axial deformation is typically negligible compared to lateral deflections under lateral loads, whilst also simplifying the analysis \citep{reese2000single, API2A}.
A Winkler foundation was chosen as it is relatively simple to implement, and it is the current method used by the OW industry, in the \textit{DNV-OS-J101} standard for the design of OW turbine structures \citep{noauthor_dnv-os-j101_2014}. Furthermore, the concept of scour can be intuitively modelled by removing springs near the surface of the soil, and reducing the effective depth of the remaining springs. Figure \ref{fig:fe_model} shows the setup of the FE model, and how it relates to the environment in which it operates.


In this initial approach, stiffness was considered as linear and not varying with foundation depth, and damping was ignored - whilst in practice this is not the case, it was considered reasonable given the primary purpose of demonstrating the methodology of the Bayesian hierarchical model. Future work does however propose to incorporate commonly-used nonlinear \textit{p-y} curve models such as those detailed in API RP 2 A-WSD \citep{API2A}. The influence of seawater on the dynamic response was modelled using an added-mass technique, as used in \citep{zuo_dynamic_2018}; this arises from the submerged body being able to impart acceleration to the surrounding fluid \citep{bi_using_2016}. The mass of additional components on the steel tower of the real structure were also accounted for, by modelling the density of the steel tower as 8500 $kg/m^3$ as in Jonkman et al. \citep{jonkman_definition_2009}. To validate the model, the first two bending-mode eigenfrequencies of the tower and nacelle section, were compared against models of the \textit{NREL} 5MW reference turbine, with the results shown in Table \ref{table:eigenfrequencies}.



\begin{table}[h]
    \small
    \begin{tabular}{|p{4.1cm}|l|l|l|l|l|}
      \hline          
      {} & {} & \multicolumn{4}{c|}{\textbf{Eigenfrequency}} \\ \hline
      \textbf{Description} & \textbf{Model} & \textbf{1st FA} & \textbf{1st SS} & \textbf{2nd FA} & \textbf{2nd SS} \\\hline
      \multirow{3}{4.1cm}{\textbf{Tower only (w/o SSI) \cite{jonkman_definition_2009}}} & {FAST} & 0.3240 & 0.3120 & 2.9003 & 2.9361 \\\cline{2-6}
                                            & {ADAMS} & 0.3195 & 0.3164 & 2.8590 & 2.9408 \\\cline{2-6}
                                            & {FE Model} & 0.3208 & 0.3208 & 2.8280 & 2.8280 \\\hline
      \multirow{4}{4.1cm}{\textbf{Tower + monopile \cite{zuo_dynamic_2018}}} & {w/o SSI} & 0.2040 & 0.2080 & 1.5620 & 1.6300 \\\cline{2-6}
                                                     & {FE model} & 0.2150 & 0.2150 & 1.5536 & 1.5536 \\\cline{2-6}
                                                     & {with SSI} & 0.1540 & 0.1560 & 1.1210 & 1.0960 \\\cline{2-6}
                                                     & {FE model} & 0.1555 & 0.1555 & 1.0481 & 1.0481 \\\hline
    \end{tabular}
    \caption{Table comparing the eigenfrequencies of models in the literature to equivalent FE models produced in this work for validation purposes. FA describes the fore-aft direction, and SS the side-to-side direction.}
    \label{table:eigenfrequencies}
  \end{table}
  
In the cases where SSI was not modelled, it can be seen that the FE model and the results shown in the literature are comparable, with a maximum difference of approximately 5\% seen in the first fore-aft eigenfrequency for the tower and monopile model; given that the blades and nacelle were approximated as lumped concentric masses in this study, this was considered acceptable. The \textit{p-y} springs were then incorporated into the FE model, with their stiffness tuned to result in natural frequencies similar to those seen in Zuo et al.~\citep{zuo_dynamic_2018} with SSI. For completeness, Table \ref{table:model_properties} details some key dimensions and properties of the main components of the model.

\begin{table}[h]
    \small
    \begin{tabular}{|l|l|l|p{4.2cm}|l|}
    \hline
    \textbf{Properties} & \textbf{Monopile} & \textbf{Tower} & \textbf{Nacelle/Rotor assembly} & \textbf{UoM} \\ \hline
    Base diameter       & 6        & 6     & 3.439                  & m           \\ \hline
    Base wall thickness & 35.1     & 35.1  & -                      & mm          \\ \hline
    Top diameter        & 6        & 3.87  & 3.439                  & m           \\ \hline
    Top wall thickness  & 35.1     & 24.7  & -                      & mm          \\ \hline
    Length              & 75       & 87.6  & 4.8                    & m           \\ \hline
    Embedded length     & 45       & -     & -                      & m           \\ \hline
    Submerged length    & 20       & -     & -                      & m           \\ \hline
    Nominal density     & 7850     & 8500  & -                     & kg/m3       \\ \hline
    Submerged density   & 8880     & -     & -                     & kg/m3       \\ \hline
    Mass                & -        & -     & 350                    & tonnes      \\ \hline
    \end{tabular}
    \caption{Properties of the FE model of the NREL 5MW reference turbines.}
    \label{table:model_properties}
\end{table}

\subsubsection{Numerical dataset generation}
\noindent In generating data for a population of wind-turbine structures, it was assumed that the expected stiffness of the foundations was allowed to vary between $K$ turbines (due to differences in foundation conditions) sampled from a global distribution. It was also assumed that foundation stiffness could vary over time (from changes in conditions) within a certain degree of variance. Given these assumptions, distributions were placed over a global stiffness expectation and variance, which were then used to sample $K$ local stiffness expectations and variances. For each $k \in K$, $N$ stiffness realisations were then drawn, which were used as inputs to the FE model to generate observations of the first bending-mode natural frequency. As a final step, these observations were corrupted by standard normal Gaussian noise, with standard deviation equal to \num{e-4}. 

\begin{figure}[h] 
\centering{\includegraphics[width=\textwidth]{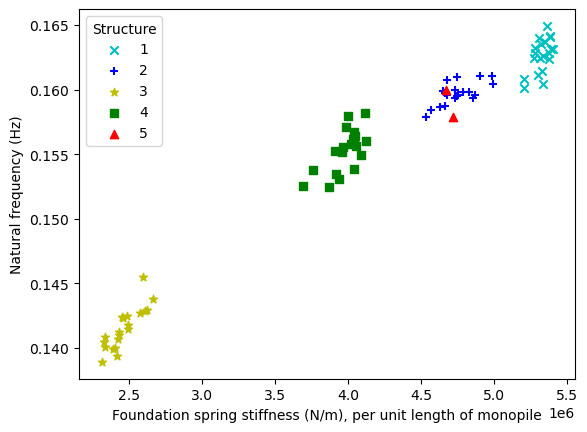}}
    \caption{Generated natural frequency observations for five turbine structures.}
\label{fig:gen_data}%
\end{figure}

In this work, to demonstrate the method, a small population of $K=5$ was used, with $N=20$ observations drawn for each $k \in K$, apart from the final structure where only two observations were drawn - this was done deliberately to create an imbalanced dataset to help demonstrate the benefits of hierarchical models. Figure \ref{fig:gen_data} shows the drawn observations used in the analyses, where the blue triangles show the fifth structure with imbalanced data.
  
\subsection{Experimental approach}\label{sec:exp_approach}
\noindent
As an alternative approach, to further demonstrate and validate the methodology, a physical experiment was designed and carried out to collect natural frequency data for a series of structures; these were designed to be similar to monopile-type wind-turbine structures from a dynamics perspective. There was an opportunity to use a wave tank, which would enable the simulation of waves to excite structures within a channeled section -- with this in mind, the following section explains the design of the experiment.



\subsubsection{Tower}

\noindent A number of key experiment design constraints and their effects on design were identified in relation to the capabilities and size of the wave tank and associated software; these were as follows:

\begin{enumerate}
\item Working water depth of wave tank of 1 m -- helps inform the length of the towers and depth of foundations
\item Working height of 1-2 m from the base of the tank -- helps inform the length of the towers
\item Optimal wave generation frequency band of 0.6-1.5 Hz -- informs target structural natural frequency for excitation
\end{enumerate}


Assuming the structure is similar to a slender, uniform cantilever beam, with a concentrated end mass; the first natural frequency can be approximated as,

\begin{equation}\label{eq:blevins}
\omega_n = \frac{1}{2\pi}\sqrt{\frac{3EI}{L^3(M + 0.24M_b)}}
\end{equation}

\noindent where $E$ is the Young's modulus, $I$ the second moment of inertia, $L$ the length of the beam, $M$ the end mass and $M_b$ the mass of the beam \cite{blevins_formulas_1979}. Given the constraints and the result of this equation, consideration was given to a range of materials including stainless steel, copper, aluminium and polymers.


  
The final design used copper piping, as it has a low stiffness relative to its density, which helped to reduce the natural frequency to within the range of the wave tank, given practical restrictions of its height. Furthermore, copper piping is cheap and readily available. For the top mass, stainless steel was used, which is relatively dense (thus requiring less volume), and it is also cheap and readily available. This mass was attached to the copper piping by a threaded bar and brazed insert. To measure the displacement of the towers over time (from which the natural frequency can be derived) laser vibrometers were used. To ensure the laser vibrometers worked effectively, small aluminium brackets were machined and attached to the top mass, which acted as flat surfaces to which reflective tape was attached. A summary of the properties and key dimensions of these components is shown in Table \ref{table:components}.

\begin{table}[h]
    \begin{tabular}{|p{3cm}|p{2.7cm}|p{2.7cm}|p{3.5cm}| }
    \hline
    \textbf{Component} & \textbf{Material} & \textbf{Dimensions (mm)} & \textbf{Total mass (inc. fixtures/fittings)} \\ 
    \hline
    tower & copper & 1500 L x 15 OD x 13.6 ID & 401 g \\ 
    \hline
    top mass & stainless steel & 78.5 L x 50.8 OD & 1280 g \\ 
    \hline
    top mass bracket & aluminium & 50 H x 50 L x 20 W & 36.4 g \\ 
    \hline
    \end{tabular}
    \caption{Key components used in the experimental model.}
    \label{table:components}
\end{table}

\subsubsection{Foundation and surrounding structure}

\noindent For the foundation material, it was decided to use a rigid foam insulation board \citep{kingspan} -- this provided a number of advantages, including (1) -- it is relatively impermeable and so it's properties are not expected to be transient, (2) because it is solid there is no loose material to foul the wave tank, and (3) it is easy to cut and manipulate, which made the simulation of scour easy to control. Separate foam boards were fixed together with adhesive to span the width of the tank, and were mounted to a steel baseplate secured to the floor of the wave tank. Because of the low density (and high buoyancy) of the foam, additional ballast was required to allow the foam and baseplate assembly to sink to the base of the tank. Circular holes were cut into the foam, in order to insert the copper piping, held in position by an interference fit; this was done in a 3 x 3 equidistant grid format, enabling multiple measurements to be gathered at once, and at a range of locations analogous to different members of a population.

To avoid a sudden discontinuity where the waves in the tank meet the foam block, a platform was constructed using steel plates, effectively raising the floor of the wave tank. This structure was designed with an initial ramped section, followed by a flat section at the height of the foam, to give a time period for the waves to stabilise before reaching the towers. This platform was attached to the tank internally using the friction from spring-loaded pads pushing against the walls of the tank. 


\subsubsection{Final design}\label{sec:final_design}

\begin{figure}[h]
    \centering{\includegraphics[width=\columnwidth]{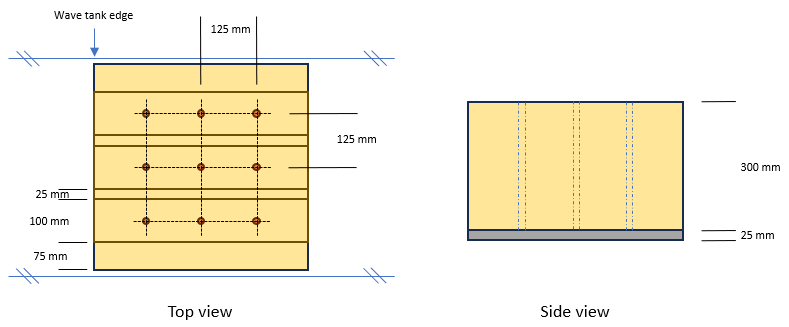}}
    \caption{Diagram of the foam foundation. Dimensions given are approximate.}
\label{fig:foundation_diagram}%
\end{figure}

\begin{figure}[h]
    \centering{\includegraphics[width=\columnwidth]{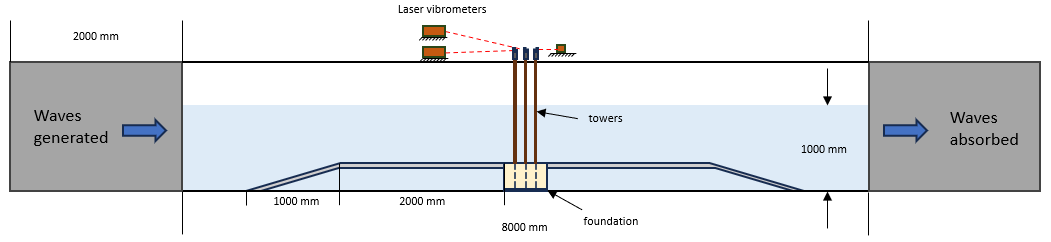}}
    \caption{Diagram of the experimental setup. Dimensions given are approximate.}
\label{fig:experiment_diagram}%
\end{figure}

\begin{figure}[h]
\centering
\begin{subfigure}{.5\textwidth}
    \centering
    \includegraphics[width=\textwidth]{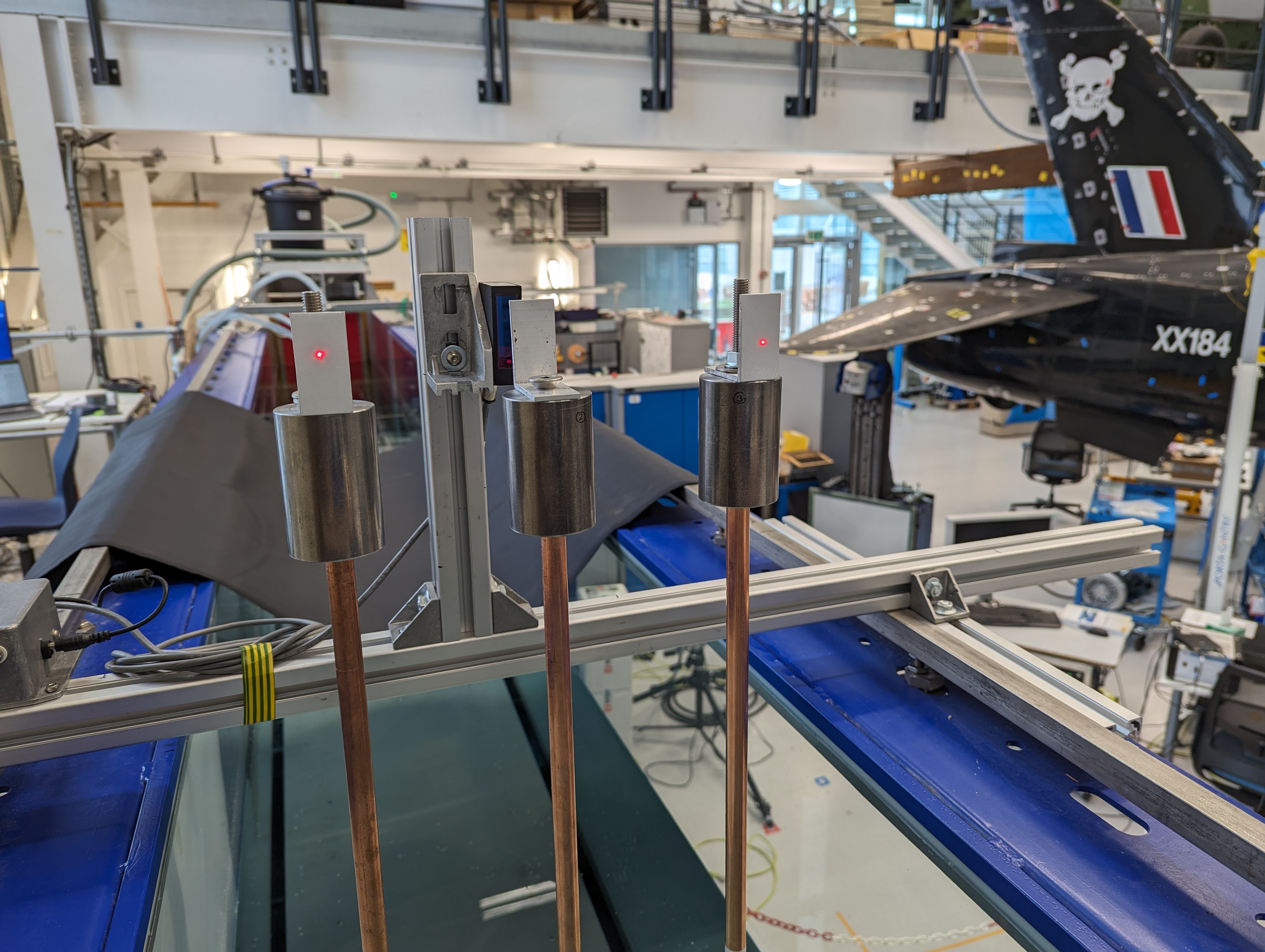}
    \caption{Top masses.}
    \label{fig:top_masses}
\end{subfigure}%
\begin{subfigure}{.5\textwidth}
    \centering
    \includegraphics[height=\textwidth]{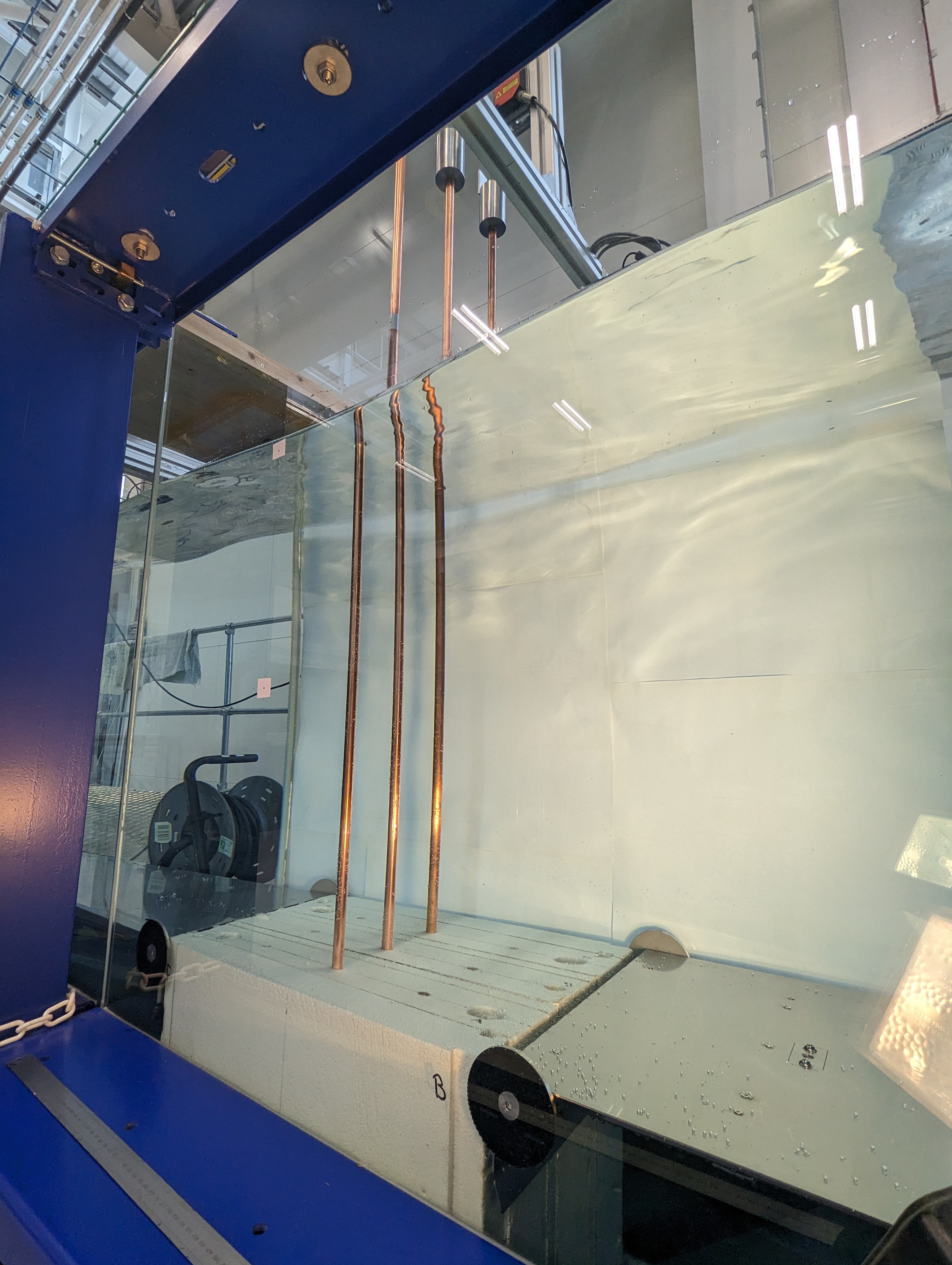}
    \caption{Underwater structure.}
    \label{fig:in_situ}
\end{subfigure}
\caption{Images showing the setup of the experiment. Image (a) shows the top masses attached to the top of the copper tubes. Image (b) shows the length of the copper tubes under the waterline and into the foundation.}
\label{fig:experiment_images}
\end{figure}

\noindent The final design of the experiment and surrounding structure is shown in Figures \ref{fig:foundation_diagram}, \ref{fig:experiment_diagram} and \ref{fig:experiment_images}.

\subsubsection{Experimental procedure}
\noindent During operation of turbines within a wind farm, benign variations in their natural frequency can be expected from factors including differences in temperature, geometrical variation and differences in the properties of seabed foundation. To simulate variation in natural frequency of the structures in the labaratory (which is otherwise carefully controlled), small additional masses were added or removed from the top mass between the experiments to change the natural frequency of the structures. Although real-world variations may be site specific, for the purposes of the experiment it was assumed to be in the order of 3-4\%. Based on equation (\ref{eq:blevins}) and a base top mass of 1.28 kg, this was deemed to require a variation in top mass of approximately 100 g. Consequently, the total top mass during experiments was allowed to vary between 1.28 - 1.38 kg, with specific masses selected to follow an approximate normal distribution centred at 1.33 kg.

Excitation during experiments was supplied by waves from the wave generator, which was programmed to use a Jonswap spectrum of waves frequencies. This spectrum was chosen to be similar to wave-loading conditions seen in the North Sea \citep{hasselmann_measurements_1973}. The Jonswap spectrum is defined as,
  
\begin{equation}\label{eq:jonswap1}
S(\omega) = \frac{\alpha g^2}{\omega^5}exp\left[-\frac{5}{4}{\left(\frac{\omega_p}{\omega}\right)}^4\right]\gamma^r
\end{equation}

\noindent where,

\begin{equation}\label{eq:jonswap2}
r = exp\left[-\frac{(\omega-\omega_p)^2}{2\sigma^{2}\omega^2_p}\right]
\end{equation}

\noindent where $\omega$ is the wave frequency (in radians per second), $\omega_p$ is the peak wave frequency (in radians per second), $\alpha$ is a constant relating to the intensity of the spectra, $g$ is the gravitational constant, $\gamma$ is the peak enhancement factor, and $\sigma$ is a constant that changes with frequency \citep{hasselmann_measurements_1973}. In the experiments, the values of the parameters were $\alpha=0.0081$, $\omega_p=0.7$ Hz, $\gamma=2$, $\sigma_a=0.07$ and $\sigma_b=0.09$. The resulting spectrum was also truncated between $0.2 - 3$ Hz in-line with the capabilities of the wave-generation machine.

In each test, three tower structures were excited and measured because of the availability of three laser vibrometers (2 $\times$ \textit{Polytec VGO-200} and 1 $\times$ \textit{Micro-epsilon OptoNCDT1300}). Tower displacements were recorded at 2048 Hz, for a minimum of 20 minutes. These time-domain displacement signals were then passed through a Butterworth filter with a passband of 0.1 - 20 Hz. Each signal was segmented into 60-second blocks with a Hanning window applied, with a 50\% overlap between each block. These blocks were then zero padded to an equivalent of 500 seconds before applying the Fast Fourier Transform (FFT) to the frequency domain data, from which the average Power Spectral Density (PSD) across all blocks was obtained. The peaks of these signals were then taken as the first natural frequencies of the towers. 
  
  

\begin{figure}[h] 
\centering{\includegraphics[width=\textwidth]{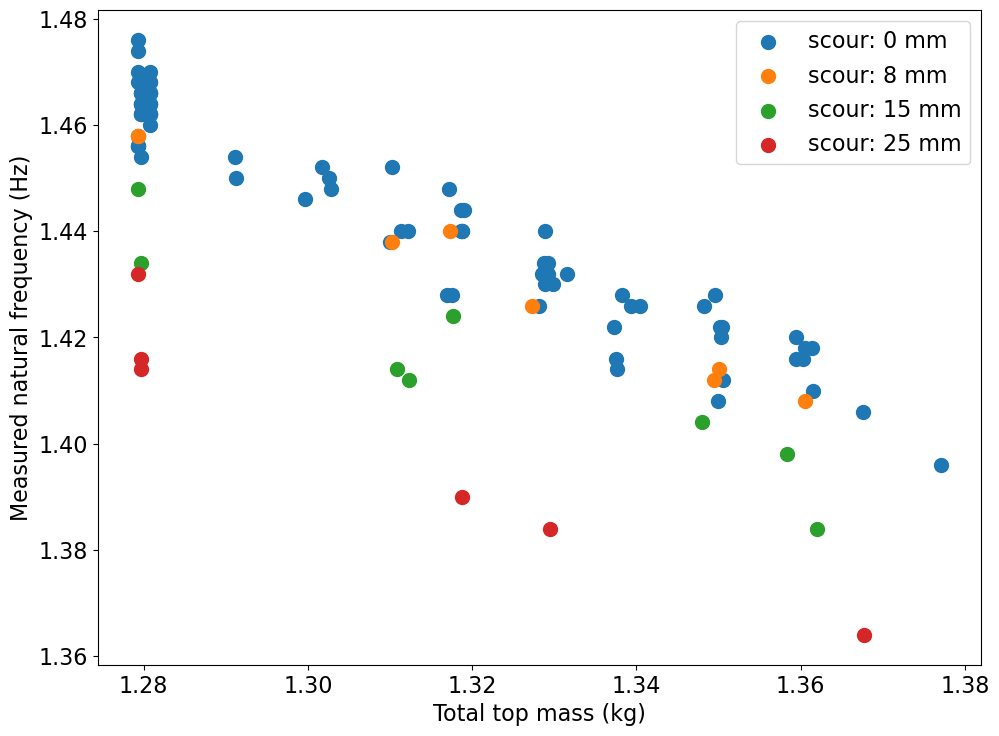}}
    \caption{Measured natural frequency measurements for nine experimental turbine structures, with varying levels of scour.}
\label{fig:exp_data}%
\end{figure}

A range of tests were completed to measure the effect of varying top masses and levels of scour on the measured natural frequency for a total of nine nominally-identical base structures; these are shown in Figure \ref{fig:exp_data}. Here, as expected, it can be seen that as the total top mass increases, the measured natural frequency decreases, for all levels of scour. Furthermore, increasing levels of scour also appear to increasingly reduce the measured natural frequency, as one would expect. For reference, as the outer diameter (D) of the tower was 15 mm, the levels of scour represent 0, 0.53, 1 and 1.66D respectively - given the overall embedment depth of 300 mm (20D), this is relatively small. The high density of data points with a total top mass of around 1.28 kg is from a number of experiments acting as reference cases.

\subsubsection{Experimental FE model}
\noindent In order to determine a functional relationship between the natural frequency measured in the experiment, and the associated foundation stiffness, an FE model was also constructed to mimic the experimental structure; this was built using the same approach as for the numerical case study, but with the dimensions and material properties as per the experimental model. 




\section{Bayesian hierarchical modelling for multitask learning}\label{sec:bayesian_hierarchical_model}

\noindent Now the datasets are established, our multitask learner (the Bayesian hierarchical model), which is used to learn the foundation stiffness parameters probabilistically is described, given the observed natural frequencies. The explanation here follows the description provided by Bull et al.~\cite{bull_hierarchical_2022}. Consider structural data, recorded from a population of $K$ similar wind-turbines in comparable soil conditions. The population data can be denoted,

\begin{equation}\label{eq:data}
\left\{\mathbf{x}_k, \mathbf{y}_k\right\}_{k=1}^K=\left\{\left\{x_{i k}, y_{i k}\right\}_{i=1}^{N_k}\right\}_{k=1}^K
\end{equation}
where $\mathbf{y}_{k}$ is a target response vector for inputs $\mathbf{x}_{k}$ and $\{x_{ik},y_{ik}\}$ are the $i^{\text{th}}$ pair of observations for turbine $k$. There are $N_{k}$ observations for each turbine and thus $\Sigma^{K}_{k=1}N_k$ observations in total. 



Models that include both shared parameters, learnt at the population level (often referred to as fixed effects), and random effects which vary between groups (or subfleets) (\textit{K}), are known as partially-pooled hierarchical models. In contrast, a no-pooling approach is where each sub-fleet has an independent model, whereby no statistical strength is shared between domains.
In the OW setting, whilst some older wind-turbines may have a rich history of data, newer wind-turbines may not, and so independent models may lead to unreliable predictions. On the other end of the spectrum, a complete-pooling approach considers all population data from a single source \citep{dardeno_hierarchical_2024}; this may lead to poor generalisation, particularly when there are significant differences between individual or groups of wind-turbines (wind farms). Hierarchical (partial-pooling) models represent a middle-ground which can be used to learn separate models for each group while encouraging task parameters to be correlated. 

In this work, the hierarchical model was used to learn global distributions over the turbine foundation stiffnesses and assumed that the stiffnesses associated with each foundation were sampled from these shared global distributions; these values were used as inputs to a pre-trained surrogate model, that mapped foundation stiffness to the natural frequency of the first bending mode using a polynomial curve fit. The surrogate model was used for efficiency of computation during the sampling, which was performed using the Monte Carlo Markov Chain (MCMC) method, via the no U-turn (NUTS) implementation of Hamiltonian Monte Carlo (HMC) \citep{homan_no-u-turn_2014}.


The following describes the setup of the model. The likelihood for the model is,
\begin{equation} \label{eq:likelihood}
\left\{\left\{\omega_{ik}\right\}_{i=1}^{N_k}\right\}_{k=1}^K {\sim} \mathrm{Normal}\left(f(s_{ik}),{\gamma}^2\right)
\end{equation}
where $\omega_{ik}$ is the $i^{th}$ observation of the first natural frequency for the $k^{th}$ turbine, $f$ is the surrogate polynomial function approximating the FE model, and $\gamma$ is the measurement error of the natural frequency. Following the Bayesian methodology, one can set prior distributions over the associated foundation stiffness realisations, $s_{ik}$ for each turbine:
\begin{equation} \label{eq:s_k}
\left\{\left\{s_{ik}\right\}_{i=1}^{N_k}\right\}_{k=1}^K \sim \mathrm{Normal}(E_{s_{k}}, V_{s_{k}})
\end{equation}

\noindent where, 

\begin{equation} \label{eq:Es}
\left\{E_{s_{k}}\right\}_{k=1}^K \sim \mathrm{Normal}(E_{\mu}, V_{\mu})
\end{equation}
\begin{equation} \label{eq:Vs}
\left\{V_{s_{k}}\right\}_{k=1}^K \sim \mathrm{Normal}(E_{\sigma}, V_{\sigma})
\end{equation}

Here, equations (\ref{eq:Es}) and (\ref{eq:Vs}) show that the turbine-level parameters for the mean and variance of the stiffness ($E_{s{_k}}$ and $V_{s{_k}}$ respectively) are normally distributed, according to the population-level parameters $E_\mu$, $V_\mu$, $E_\sigma$ and $V_\sigma$. As each of these population-level parameters (as well as the noise parameter $\gamma$) are also uncertain, prior distributions were also placed over them (known as hyper-priors). These hyper-priors give an opportunity for domain expertise to be encoded into the model; in this case, one knows that both the stiffness values ($s_{i_k}$) and all parameters associated with variance must be strictly positive. To ensure positivity, gamma distribution hyper-priors were used for all population-level parameters.







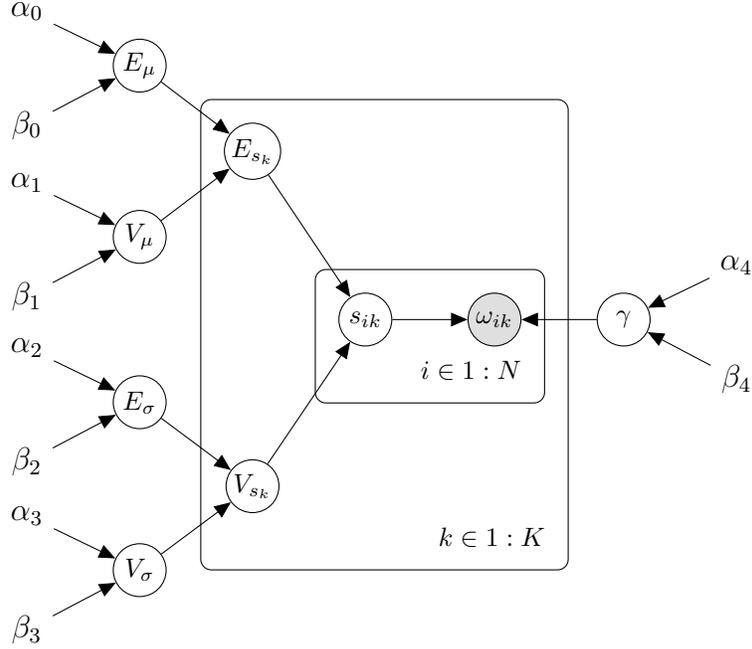
\begin{figure}[t]
    \centering
    \begin{tikzpicture}
    \node[obs]                              (w) {$\omega_{ik}$};
    \node[latent, left=1cm of w] (s) {$s_{ik}$};
    \node[latent, above=1.5cm of s, xshift=-1.5cm] (E_s) {$E_{s_{k}}$};
    \node[latent, below=1.5cm of s, xshift=-1.5cm] (V_s) {$V_{s_{k}}$};
    \node[latent, above=0.4cm of E_s, xshift=-1.5cm] (E_mu) {$E_{\mu}$};
    \node[latent, below=0.4cm of E_s, xshift=-1.5cm] (V_mu) {$V_{\mu}$};
    \node[latent, above=0.4cm of V_s, xshift=-1.5cm] (E_sig) {$E_{\sigma}$};
    \node[latent, below=0.4cm of V_s, xshift=-1.5cm] (V_sig) {$V_{\sigma}$};
    \node[latent, right=1cm of w] (gamma) {$\gamma$};
    \node[above=0.1cm of E_mu, xshift=-1.5cm] (E_mu0) {$\alpha_0$};
    \node[below=0.1cm of E_mu, xshift=-1.5cm] (E_mu1) {$\beta_0$};
    \node[above=0.1cm of V_mu, xshift=-1.5cm] (V_mu0) {$\alpha_1$};
    \node[below=0.1cm of V_mu, xshift=-1.5cm] (V_mu1) {$\beta_1$};

    \node[above=0.1cm of E_sig, xshift=-1.5cm] (E_sig0) {$\alpha_2$};
    \node[below=0.1cm of E_sig, xshift=-1.5cm] (E_sig1) {$\beta_2$};
    \node[above=0.1cm of V_sig, xshift=-1.5cm] (V_sig0) {$\alpha_3$};
    \node[below=0.1cm of V_sig, xshift=-1.5cm] (V_sig1) {$\beta_3$};

    \node[above=0.1cm of gamma, xshift=1.5cm] (gamma0) {$\alpha_4$};
    \node[below=0.1cm of gamma, xshift=1.5cm] (gamma1) {$\beta_4$};

    \edge {s,gamma} {w} ; %
    \edge {E_s,V_s} {s} ; %
    \edge {E_mu,V_mu} {E_s} ; %
    \edge {E_sig,V_sig} {V_s} ; %

    \edge {E_mu0, E_mu1} {E_mu} ; %
    \edge {V_mu0, V_mu1} {V_mu} ; %
    \edge {E_sig0, E_sig1} {E_sig} ; %
    \edge {V_sig0, V_sig1} {V_sig} ; %
    \edge {gamma0, gamma1} {gamma} ; %

    \plate [inner sep = .3cm]{N} {(w)(s)} {$i\in1:N$} ;
    \plate [inner sep = .3cm]{k} {(w)(s)(E_s)(V_s)(N.north west)(N.south east)} {$k\in1:K$} ;

    \end{tikzpicture}
    \caption{A graphical model representing the hierarchical model with partial pooling.}
    \label{fig:hierarchical_model}
\end{figure}

A graphical model depicting the hierarchical structure of the model can be seen in Figure \ref{fig:hierarchical_model}. Latent and observed variables are depicted as unshaded and shaded circled nodes respectively, with the plates indicating multiple instances of their contained nodes. The un-circled parameters are the constants used in the hyper-prior distributions.

\subsection{Training procedure}
\noindent As is typical, because of the high computational demand of the FE models, it was infeasible to incorporate them directly as the mean function within the probabilistic models. Instead, the surrogate polynomial functions were used that closely matched the relationship between the input foundation spring stiffness per unit length, and the natural frequency of the first bending mode. In this case, the functions were fifth-order.
The MCMC sampling was then carried out using four chains, each with 2000 warm-up samples, followed by a further 2000 samples. The warm-up samples were discarded to diminish the influence of starting values \citep{gelman_bayesian_2013}.


\section{Results and Discussion}
\subsection{Numerical foundation model}

\begin{figure}[h] 
\centering{\includegraphics[width=\columnwidth]{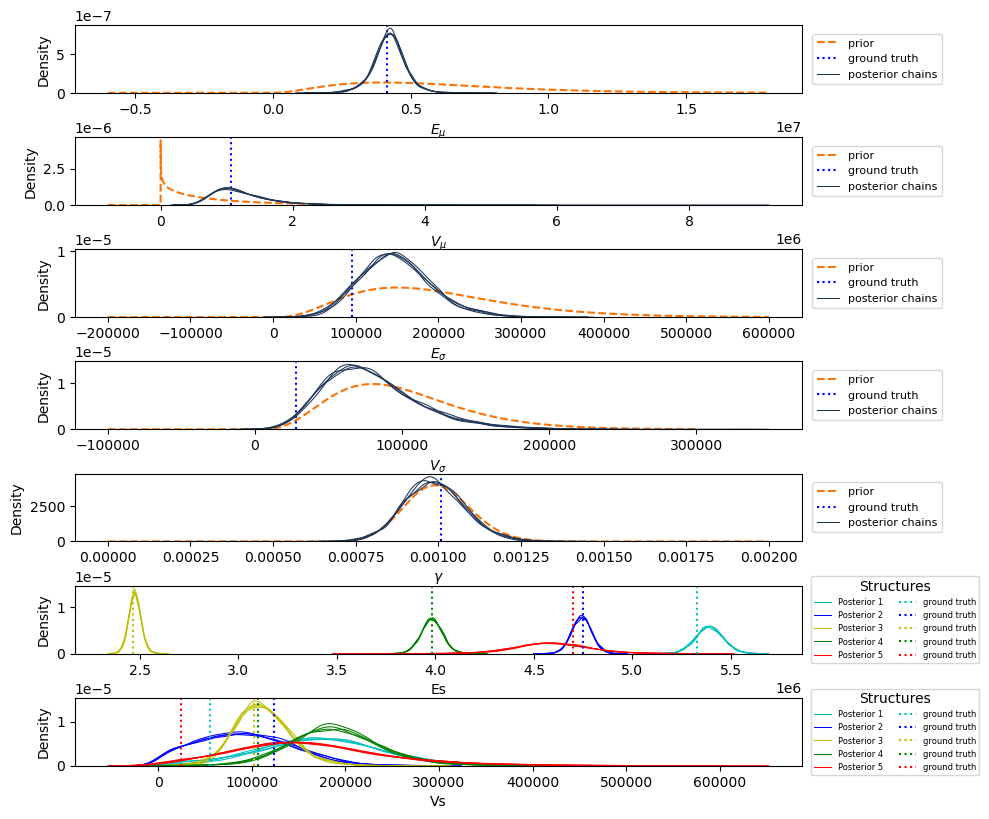}}
    \caption{Plots showing the density of the posterior samples for each chain, for the learned parameters of the NREL 5MW model. Vertical dotted lines indicate the ground truth expected values of the generated data to be learned. Dashed orange lines show the prior distributions applied to the global parameters.}
\label{fig:base_post}%
\end{figure}

\noindent Density plots for the 2000 posterior samples for each chain are shown in Figure~\ref{fig:base_post}, representing the inferred distributions of the parameters. The vertical dashed lines mark the true expected values of the distributions used in generating the data. 
As previously discussed, the hyper-priors for the population-level parameters follow gamma distributions to ensure positivity - these are shown as dashed orange lines in the figure.

During testing, one issue with this model design that was identified, was that both the $\gamma$ and $E_\sigma$ parameters can be used to explain variance in the natural frequency observations, which resulted in model convergence issues and their posterior samples being negatively correlated. To help alleviate this problem, a relatively precise and narrow prior was placed over the $\gamma$ parameter - which, given that this represents measurement noise which is likely well known for a given measurement device, is not unrealistic. The resulting model had all \textit{$\hat{R}$} convergence statistics approximately equal to one, a useful metric for assessing convergence \cite{vehtari_rank-normalization_2021}.

For the first four population-level parameters $E_\mu$, $V_\mu$, $E_\sigma$ and $V_\sigma$, relative to their comparatively-vague prior distributions it can be seen that the modes of the posterior densities have shifted towards their ground truth expected values, while their variances have also reduced; this is as expected. It can be seen that the posterior density for the $\gamma$ parameter is very similar to the prior density; it is believed that the shallow tails of the prior distribution constrain the MCMC algorithm from exploring more widely, and thus any additional variance seen in the observations is explainable via the $E_\sigma$ parameter, which is as desired.

For the turbine-level parameter $E_s$, the modes of all posterior densities are close to their expected values, while the variance of the fifth structure is significantly higher as it has only two data points compared to 20 for the other structures. Furthermore, for this fifth structure, the posterior density of $E_s$ is biased towards the left of the ground truth expected value, which is indicative of it being pulled towards the global expected mean ($E_\mu$) of approximately 4e6; this shows the benefit of hierarchical models with partial pooling, whereby data-poor domains benefit from the statistical significance of the population-level parameters. Similar observations can be made for parameter $V_s$. These observations give confidence that the algorithm is working as intended, where, based on observations of variable natural frequency from a numerical structure, probability density functions of the associated foundation stiffness can be obtained for individuals and populations of structures.


%


\subsection{Experimental foundation model}

\begin{figure}[h] 
\centering{\includegraphics[width=\textwidth]{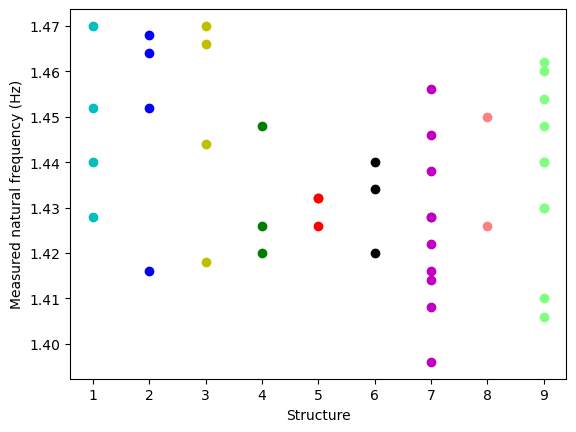}}
    \caption{Plot showing the experimental observations of natural frequency, for each of the nine structures tested.}
\label{fig:exp_data_model}%
\end{figure}

\noindent Having demonstrated the method in a numerical study, it is now applied to the experimental data. The first set of results, use only the data points that are not affected by scour. Figure \ref{fig:exp_data_model} shows this subset of observations, for each population member.

\begin{figure}[h] 
\centering{\includegraphics[width=0.9\columnwidth]{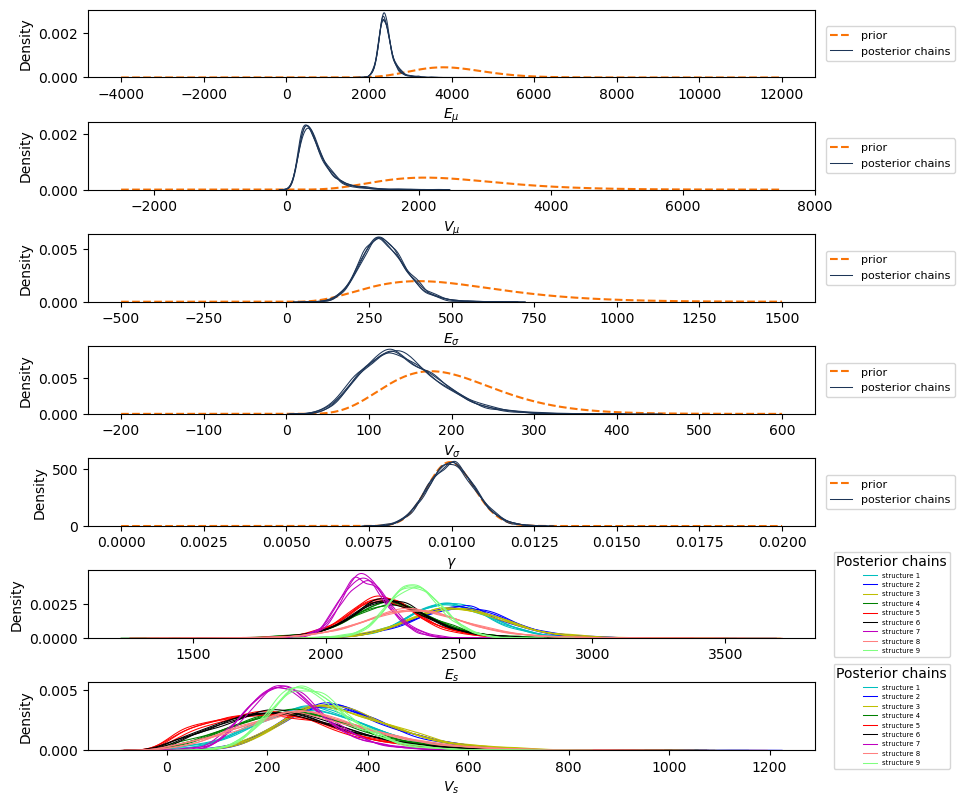}}
    \caption{Plots showing the density of the posterior samples for each chain, for the learned parameters of the experimental model. Dashed (orange) lines show the prior distributions applied to the global parameters.}
\label{fig:lvv_post}%
\end{figure}

Relatively-broad priors were used for the population-level parameters, apart from the noise parameter $\gamma$ which again was assumed to be relatively well known, as it directly relates to sensor accuracy. Figure \ref{fig:lvv_post} shows the posterior densities of the learned parameters.

As with the numerical foundation model, the posterior densities of the population-level parameters (apart from $\gamma$) show a significant reduction in variance, with their modes moving away from the prior modes, indicative that the model has learned from the observations. For the turbine-level parameters $E_s$ and $V_s$, the posterior variances appear tighter for structures with more data (e.g.~Seven and Nine), and wider for structures with fewer data (e.g.~Four, Five and Eight); this matches intuition. Finally, the parameter $\gamma$ behaves similarly, whereby the posterior density closely matches the prior density.

If the approach were to be applied to real wind-turbine dynamic data, this could give greater confidence in foundation-stiffness parameters for a given area or for comparable ground conditions. This reduced uncertainty, could help to minimise the risk of excessive fatigue damage by helping designers avoid structural natural frequencies being too close to wind and wave loading conditions. Furthermore, the model results also provide a basis for anomaly detection (such as resulting from scour), as described in the next section.

\subsection{Anomaly/Scour Detection}\label{sec:scour}

\begin{figure}[h]
\centering
\begin{subfigure}{.4\textwidth}
    \centering
    \includegraphics[width=\textwidth]{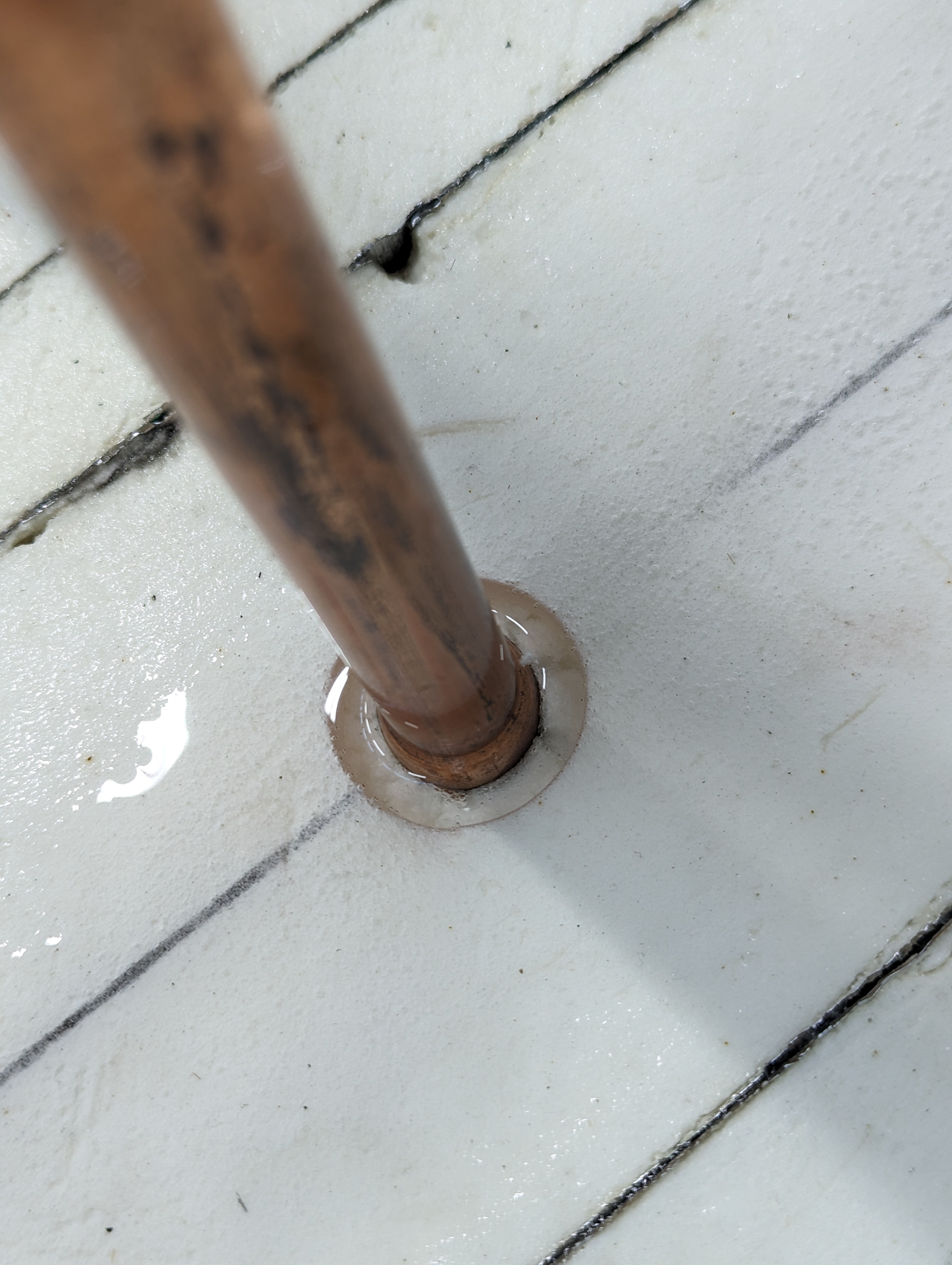}
    \caption{Example of scoured material around the base of the tower.}
    \label{fig:scour_hole}
\end{subfigure}%
\hspace{0.05\textwidth} 
\begin{subfigure}{.4\textwidth}
    \centering
    \includegraphics[width=\textwidth]{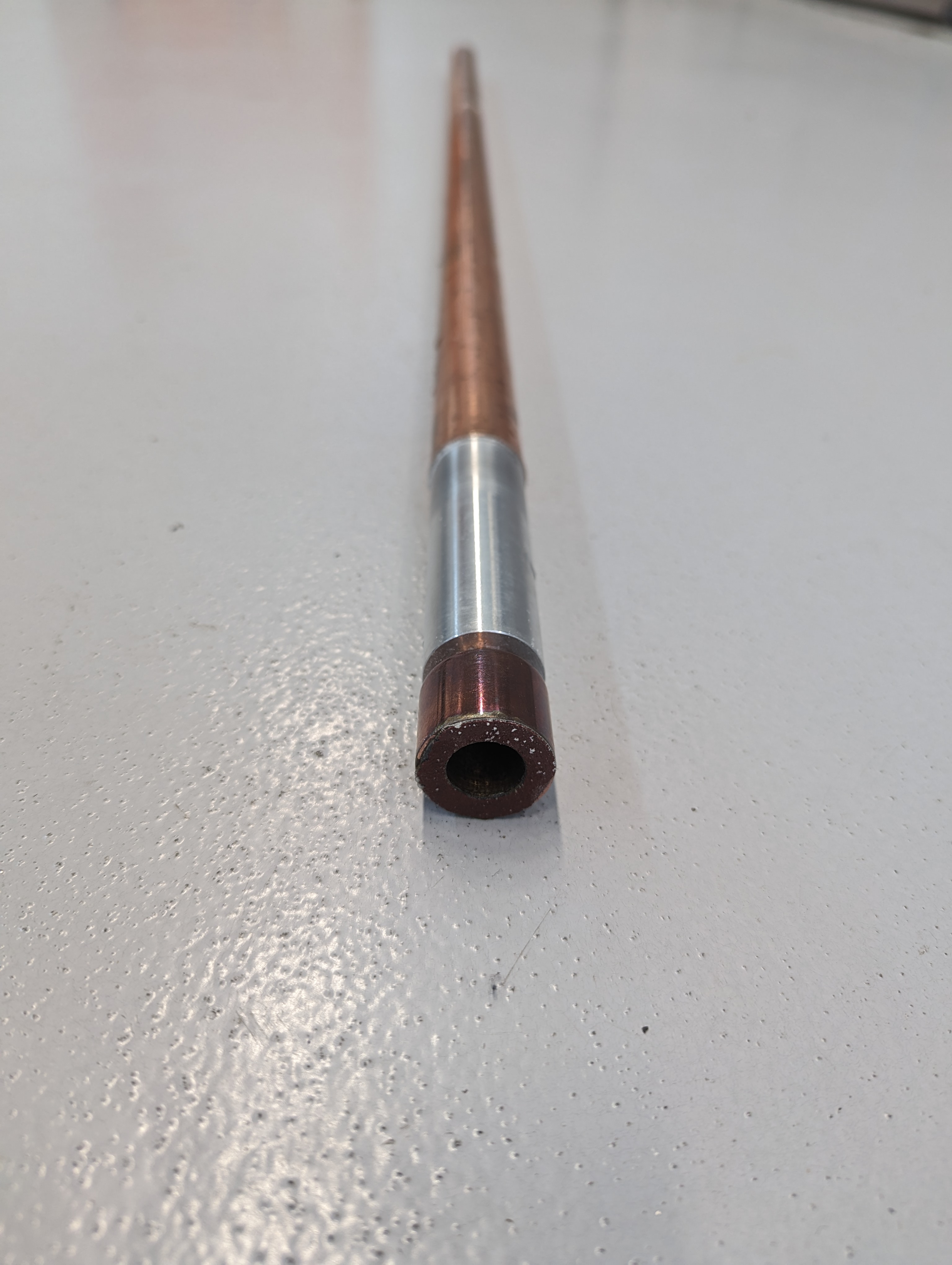}
    \caption{Abrasion tool used to create the scour hole.}
    \label{fig:scour_tool}
\end{subfigure}
\caption{Images showing how scour was performed in the experiments. Image (a) shows a hole cut into the foam using the abrasion tool (shown in image (b)).}
\label{fig:scour_method}
\end{figure}

For some of the tests during the experiment, scour was introduced by removing material at the top of the foundation--an example of this is shown in Figure \ref{fig:scour_hole} where material was removed using abrasion via the tool shown in Figure \ref{fig:scour_tool}.

\begin{figure}[h] 
\centering{\includegraphics[width=\textwidth]{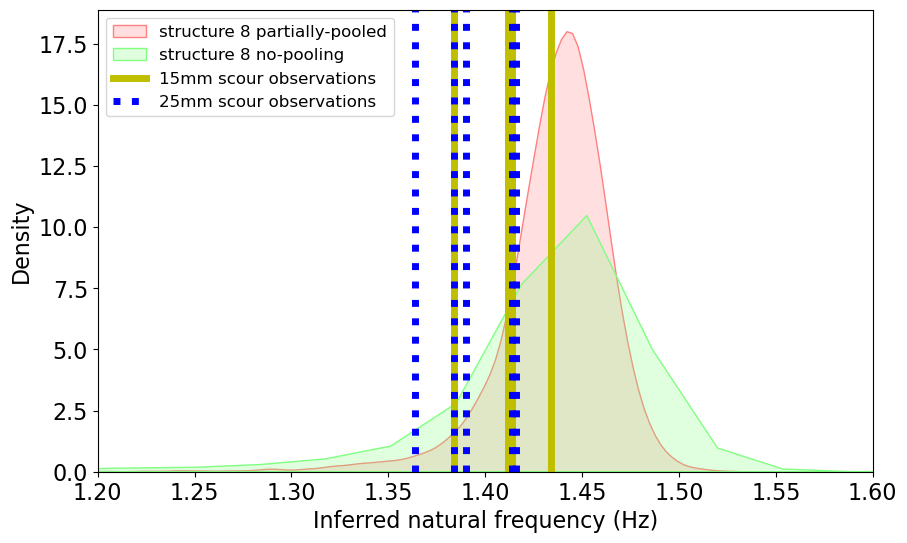}}
    \caption{Figure showing the partially-pooled and no-pooling posterior predictive distributions of the natural frequency for the $8^{th}$ structure, in relation to new observations which are influenced by scour. The solid yellow lines represent observations with 15 mm of scour, while the dotted blue lines represent observations with 25 mm of scour.}
\label{fig:scour_posterior}%
\end{figure}

As an example of how the posterior distributions of the stiffness could be used for anomaly detection, and to demonstrate the benefit of a partial-pooling over no-pooling, Figure \ref{fig:scour_posterior} shows the associated posterior predictive distributions for the natural frequency for the eighth structure. These posteriors were computed by drawing a stiffness for each MCMC sample as per equation \ref{eq:s_k}, applying the surrogate model and drawing a natural frequency as per the likelihood in equation \ref{eq:likelihood}.
The vertical lines represent additional observations of natural frequency, but with scour depths of 15 mm (solid yellow line) and 25 mm (dotted blue line); given the 15 mm diameter (D) of the towers this is equivalent to 1 and 1.67 D.

Here it can be seen that the presence of scour appears to result in natural frequencies lower than the modes of the posterior distributions, with deeper scour (25 mm) resulting in a greater shift, as one would expect. In each case, the probabilities of natural frequencies equal to or less than the observations were calculated (using the posterior predictive samples calculated for Figure \ref{fig:scour_posterior}), and are displayed in Table \ref{table:scour_obs}. For all but the first scour observation (which by chance was close to the mode of both distributions), significant reductions in the associated probabilities for the partially-pooled approach can be seen, from about 28\% for observation five to 70\% for observation six. These reduced probabilities for the partially-pooled approach demonstrate the benefit of the hierarchical model, which can more confidently identify that the new observations are outliers, and that a step change in the underlying dynamic properties in the structure has been observed.
In practice, it may be prudent to compare these probabilities against pre-determined thresholds to assess novelty for consistency.




\begin{table}[h]
    \small
    \begin{tabular}{|l|p{2.3cm}|p{2cm}|p{2.5cm}|p{3.5cm}|}
    \hline
    \textbf{Obs.} & \textbf{Scour depth (mm)} & \textbf{Nat. freq. (Hz)} & \textbf{Prob. No-pooling} & \textbf{Prob. Partially-pooled}\\ \hline
        1 & 15 & 1.434 & 0.395 & 0.398 \\\hline
        2 & 15 & 1.414 & 0.219 & 0.152 \\\hline
        3 & 15 & 1.384 & 0.110 & 0.043 \\\hline
        4 & 15 & 1.412 & 0.209 & 0.136 \\\hline
        5 & 25 & 1.416 & 0.232 & 0.166 \\\hline
        6 & 25 & 1.364 & 0.078 & 0.023 \\\hline
        7 & 25 & 1.384 & 0.110 & 0.043 \\\hline
        8 & 25 & 1.390 & 0.123 & 0.054 \\\hline
        9 & 25 & 1.414 & 0.219 & 0.152 \\\hline
    \end{tabular}
    \caption{Probabilities of the natural frequencies (influenced by scour) equal to or less than the observations, given the posterior predictive natural frequencies for structure 8.}
    \label{table:scour_obs}
\end{table}



\section{Conclusions and further work}

\noindent A Bayesian hierarchical model and associated surrogate FE models were used in two separate case studies to model the uncertainty of foundation stiffnesses of wind-turbines. The first of these was a numerical case study using an FE model of the \textit{NREL} 5 MW reference turbine \citep{jonkman_definition_2009}, to generate first bending mode natural frequency observations. In the second case study, natural frequency observations were instead created from measurements of the dynamics of a scaled down turbine-like structures in a physical wave tank. In both cases, the method shows that \textit{a priori} beliefs on uncertainty of foundation stiffness can be updated based on observations of natural frequency (via the surrogate FE model), at both an individual structural level, and at a population level. For data-poor domains (or structures), the hierarchical approach allows data-poor domains to leverage statistical weight from the population level, helping to keep uncertainty relatively low. The refined posterior uncertainties on parameters could help inform future structural design, such as in nearby locations and/or in areas with comparable soil conditions. Furthermore, they may be used as a basis for more robust anomaly detection, such as in detecting scour around the foundation of monopiles.

While the focus of this work was on demonstrating the Bayesian hierarchical model, in order to develop this approach for real-world use, the foundation stiffness should be modelled as nonlinear, which is highly likely to be the case in practice. Nonlinearity could be achieved by incorporating widely-used models such as those for sand or clay in \citep{API2A}. Furthermore, sources of uncertainty such as the magnitude of wind and wave loading and the effect of damping in the foundation (likely to be significant) could be considered.

\section{Data availability}
For the purpose of reproducibility, all data and code used in this work has been made available online at: \url{https://github.com/smbrealy/foundation_modelling} and \url{https://github.com/smbrealy/hierarchical_foundations}.

\section{Acknowledgements}
The author(s) gratefully acknowledge Michael J. Dutchman who contributed significant time and expertise to the design and construction of the physical experiments - which without it would have not been possible to get the results.

The authors also gratefully acknowledge the support of the UK Engineering and Physical 
Sciences Research Council (EPSRC) and the Natural Environment Research Council (NERC), 
via grant references EP/W005816/1, EP/R003645/1, and EP/S023763/1. For the purpose 
of open access, the author(s) have applied a Creative Commons Attribution (CC BY) 
licence to any Author Accepted Manuscript version arising.



\bibliographystyle{elsarticle-num-names} 
\bibliography{cas-refs}





\end{document}